\documentclass[10pt,twocolumn,letterpaper]{article}

\usepackage{cvpr}              

\usepackage{graphicx}
\usepackage{amsmath}
\usepackage{amssymb}
\usepackage{booktabs}

\usepackage{amsthm,amsmath,amssymb}

\usepackage{mathrsfs}
\usepackage[pagebackref,breaklinks,colorlinks]{hyperref}
\def\degree{${}^{\circ}$}
\usepackage{multirow}
\usepackage[capitalize]{cleveref}
\crefname{section}{Sec.}{Secs.}
\Crefname{section}{Section}{Sections}
\Crefname{table}{Table}{Tables}
\crefname{table}{Tab.}{Tabs.}

\def\cvprPaperID{2994} 
\def\confName{CVPR}
\def\confYear{2022}

\begin{document}

\title{ACR-Pose: Adversarial Canonical Representation Reconstruction Network for Category Level 6D Object Pose Estimation}

\author{\textbf{Zhaoxin Fan$^1$, Zhengbo Song$^3$,  Jian Xu$^4$, Zhicheng Wang$^4$, Kejian Wu$^4$, Hongyan Liu$^2$, and Jun He$^1$}\\
Renmin University of China, Tsinghua University\\
Nanjing University of Science and Technology, Nreal\\
{\tt\small \{fanzhaoxin,hejun\}@ruc.edu.cn, \{xujian,zcwang,kejian\}@nreal.ai}\\
\tt\small {hyliu@tsinghua.edu.cn, songzb@njust.edu.cn}
}
\maketitle

\begin{abstract}
  Recently, category-level 6D object pose estimation has achieved significant improvements with the development of reconstructing canonical 3D representations. However, the reconstruction quality of existing methods is still far from excellent.  In this paper, we propose a novel \textbf{A}dversarial \textbf{C}anonical Representation \textbf{R}econstruction Network named \textbf{ACR-Pose}. ACR-Pose consists of a Reconstructor and a Discriminator. The Reconstructor is primarily composed of two novel sub-modules: Pose-Irrelevant Module~(PIM) and Relational Reconstruction Module~(RRM). PIM tends to learn canonical-related features to make the Reconstructor insensitive to rotation and translation, while RRM explores essential relational information between different input modalities to generate high-quality features. Subsequently, a Discriminator is employed to guide the Reconstructor to generate realistic canonical representations. The Reconstructor and the Discriminator learn to optimize through adversarial training. Experimental results on the prevalent NOCS-CAMERA and NOCS-REAL datasets demonstrate that our method achieves state-of-the-art performance.
\end{abstract}

\section{Introduction}
\label{sec:intro}

6D object pose estimation is at the core of many applications, such as robotic grasping \cite{bicchi2000robotic,saxena2008robotic,tremblay2018deep}, augmented reality \cite{azuma1997survey,carmigniani2011augmented,vavra2017recent}, and autonomous driving  \cite{song2019apollocar3d,ke2020gsnet,sun2020scalability}. Given an observation (RGB, RGBD, or LiDAR scan), 6D object pose estimation aims to estimate the rotation and translation of the target object in the camera coordinate system. Over the last decade, numerous 6D object pose estimation works\cite{xiang2017posecnn,tekin2018real,peng2019pvnet,li2018deepim,zakharov2019dpod,song2020hybridpose,he2020pvn3d} have emerged and been deployed in industrial products, demonstrating the usefulness of this line of research. However, most of the existing methods are instance-level approaches, i.e., a well-trained model works only for a particular object.
This prohibits large-scale applications in the wild, as it is both memory-inefficient and labor exhausting to train and deploy models per object.

\begin{figure}[t]
  \centering
   \includegraphics[width=0.95\linewidth]{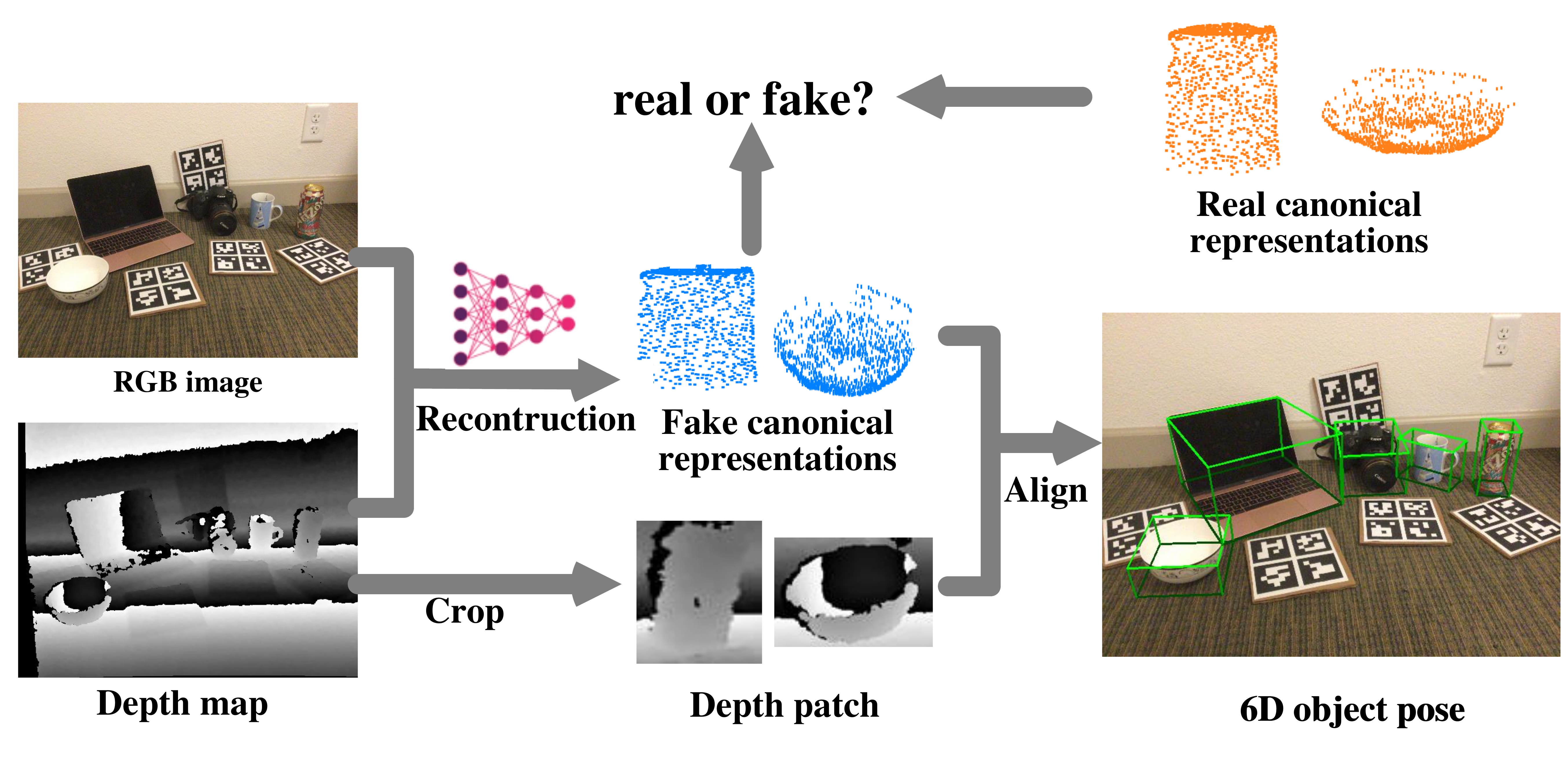}
   \vspace{-0.1in}
   \caption{Our Adversarial Canonical Representation Reconstruction Network for reconstructing canonical representations, which are then aligned with the object depth observation to estimate the 6D object pose.}
   \label{fig:pipeline}
   \vspace{-0.1in}
\end{figure}

To address this issue, category-level 6D object pose estimation has become increasingly popular recently among academia and industry. Given a RGBD observation as input, a deep learning network is often used to predict a canonical representation \cite{wang2019normalized} of the target object. The canonical representation refers to a normalized 3D model that integrates a category of objects' most representative characteristics, independent of the object's pose, such as the NOCS \cite{wang2019normalized} and the CASS \cite{chen2020learning} representation. Once the canonical representation is obtained, it can be used to align with the back-projected object depth observation to recover the object's pose. Therefore, the key to successful category-level object pose estimation is to accurately reconstruct the canonical representation.

Although existing methods \cite{wang2019normalized,chen2020learning,tian2020shape,chen2021fs,lin2021dualposenet} have achieved promising 6D object pose estimation performance through the canonical representation, their reconstruction quality still needs improvement. Specifically, previous methods have the following drawbacks: 1) They are ineffective in learning canonical-related features due to their sensitivity to pose-related characteristics. 2) They ignore the critical relational information between different input modalities.  3) Some of their reconstruction results appear to be unrealistic.

In this paper, we propose an \textbf{A}dversarial \textbf{C}anonical Representation \textbf{R}econstruction pipeline called \textbf{ACR-Pose}, which aims at reconstructing high quality canonical representations for precise 6D object estimation as shown in Fig.~\ref{fig:pipeline}. In ACR-Pose, we first detect and segment out objects from the RGBD images. Then, taking the image patch and depth region of an object as input, we train a Reconstructor and a Discriminator in an adversarial manner to generate realistic canonical representations.

Specifically, as the Reconstructor plays a significant role in generating high-quality canonical representations, two novel sub-modules are proposed to learn more discriminative and robust features within the Reconstructor. First, we note that observations of the objects contain both pose-related information~(rotation and translation) and shape information, and existing canonical representation reconstruction models \cite{wang2019normalized,tian2020shape} are sensitive to pose-related characteristics. However, since the goal of reconstruction is \emph{canonical}, the rotational and translational cues in the object observation must be filtered out, while only the shape information should be preserved. To achieve this, we introduce a Pose-Irrelevant Module (PIM) so as to learn canonical-related features only.
Second, inspired by \cite{tian2020shape}, we take a learned shape prior as an input of our network, in addition to the RGB and depth images. We hypothesize that the relational information between these three sources of information is important for high-quality reconstruction. Therefore, a Relational Reconstruction Module (RRM) is introduced to learn this relational information of the three inputs.
Finally, utilizing the features learned from the RRM, the canonical representation can be reconstructed using a Shape Prior Deformation step \cite{tian2020shape}. Afterwards, 6D object pose can be easily solved by aligning the canonical representation with the back-projected object depth using the Umeyama algorithm \cite{umeyama1991least}.

Furthermore, we find that the reconstruction results could be unrealistic in some cases, which may cause inaccurate alignment results. To resolve this issue, we adopt the idea of Generative Adversarial Network (GAN) \cite{zhao2016energy,zhu2017unpaired} in our pipeline and propose to use a Discriminator to adversarially guide the Reconstructor, making it more creative so that it can produce more realistic canonical representations. As a result, experiments on the NOCS-CAMERA and NOCS-REAL datasets show that our ACR-Pose can achieve state-of-the-art performance in both synthetic and real scenarios.

To summarize, our contributions are:

1) We propose ACR-Pose, an Adversarial Canonical Representation Reconstruction Network, for learning high-quality canonical representations trained in an adversarial manner. To the best of our knowledge, we are the first to introduce adversarial training for category-level 6D object pose estimation.

2) We propose a Pose-Irrelevant Module and a Relational Reconstruction Module to form the ACR-Pose's Reconstructor, through which high-quality canonical-related features and relational features are effectively learned for credible reconstruction.

3) We conduct extensive experiments to evaluate our model and compare it with previous works, where our model achieves state-of-the-art performance on both synthetic and real world datasets.

\begin{figure*}[t]
  \centering
   \includegraphics[width=0.98\linewidth]{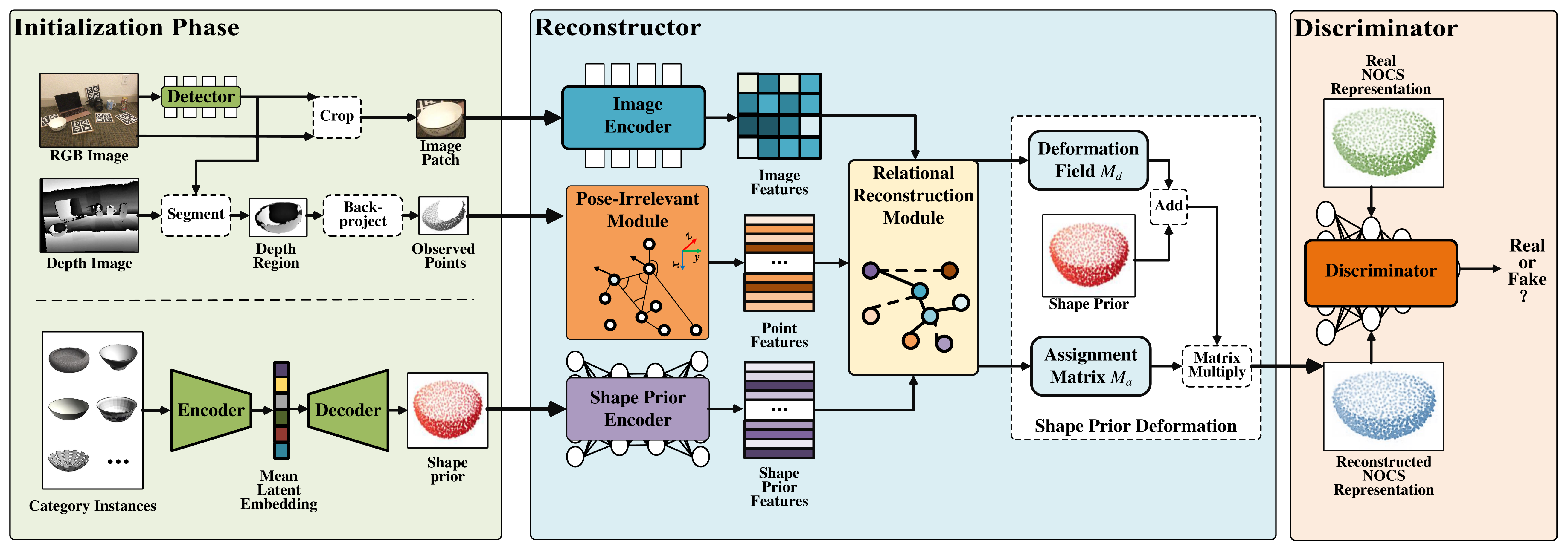}
   \vspace{-0.1in}
   \caption{The pipeline of ACR-Pose. We use the image patch, depth region, and shape prior as inputs at Initialization Phase. We predict the NOCS representation using our Reconstructor. We adopt adversarial training using the Discriminator.}
   \label{fig:trirpose}
    \vspace{-0.1in}
\end{figure*}

\section{Related Work}
In this section, we briefly review recent methods in 6D object pose estimation. Since our model is a deep learning model, we mainly introduce the deep learning-based counterparts. Then, we present works related to adversarial training.

\subsection{6D Object Pose Estimation}
Instance-level 6D object pose estimation only estimates the 6D pose of a particular object can be divided into five parts: direct-methods \cite{xiang2017posecnn,kehl2017ssd,zhang2021efficientpose}, keypoint-based methods \cite{peng2019pvnet,song2020hybridpose,tekin2018real}, dense coordinate-based methods \cite{li2019cdpn,park2019pix2pose,zakharov2019dpod}, refinement based methods \cite{li2018deepim,manhardt2018deep,yen2020inerf} and self-supervised methods \cite{wang2020self6d,sock2020introducing}. There are also many methods propose to utilize RGBD data as input for instance-level object pose estimation \cite{he2020pvn3d,wu2021vote,he2021ffb6d,rusinkiewicz2001efficient}.   For more details, we refer readers to Fan et al. \cite{fan2021deep} for a comprehensive overview. Though promising, generalization of instance-level methods is limited because we have to train different models for different objects even though they belong to the same category. This requires huge computational costs at both training and inference time. In this paper, we explore category-level 6D object pose estimation owing to its better generalization ability towards different objects.

NOCS \cite{wang2019normalized} is the first deep learning-based category-level method. It proposes the Normalized Object Coordinate Space (NOCS) representation to handle huge shape variance between different objects. Then, Chen et al. \cite{chen2020learning} propose another canonical representation named Canonical Shape Space (CASS). Both NOCS and CASS attract the interests of the following works. For example, SPD \cite{tian2020shape} proposes a shape prior deformation method for accurate NOCS representation reconstruction utilizing a category-prior as an additional model input. Lee et al. \cite{lee2021category} propose to simultaneously predict the NOCS representation and the metric-scale object shape from the RGB observation for category-level object pose detection, trying to remove the dependence on the depth information. However, its performance still lags behind RGBD based methods. Recently, DualPoseNet \cite{lin2021dualposenet} proposes a Dual-Pose Network to learn pose consistency to improve the pose estimation accuracy, and FS-Net \cite{chen2021fs} introduces a novel data augmentation method for improving models' performance. Nevertheless, no matter what efforts they make to improve performance, most of the existing methods adopt  the idea of reconstructing the canonical representation in their models.

Our method also reconstructs the NOCS representation for category-level 6D object pose estimation. We argue that the potential of the NOCS representation has not been fully explored. Therefore, we propose ACR-Pose to improve the reconstruction quality through adversarial training, which will be presented in detail later.

\subsection{Generative Adversarial Networks (GANs)}
Generative Adversarial Network (GAN) \cite{goodfellow2014generative,mirza2014conditional} is widely used in many generation tasks like style translation \cite{karras2019style}, image generation \cite{liu2016coupled}, human reconstruction \cite{kocabas2020vibe}, etc. Usually, GAN consists of a Generator and a Discriminator. The former is the main branch to predict the task goal, while the latter is used to distinguish real sample (ground-truth) and fake sample (output of the generator). The key insight of GAN is to use the Discriminator to guide the Generator through adversarial training. CGAN \cite{goodfellow2014generative} introduces the conditional version of GANs, which makes the generation process of GANs being controlled.  DCGANs \cite{yu2017unsupervised} proposes to use GANs for unsupervised learning with CNNs. WGAN \cite{arjovsky2017wasserstein} improves the stability of learning, getting rid of problems like mode collapse. CycleGAN \cite{zhu2017unpaired} proposes the concept of utilizing cycle consistency for training GANs in an unsupervised manner. Recently, there are aslo many works utilizing GANs for 3D vision tasks. For example, Li et al. \cite{li2019pu} introduces PUGAN for point cloud up-sampling. Zhang et al. \cite{zhang2021unsupervised} introduces an unsupervised point cloud completion method mainly benefits from GAN inversion. There also exists methods like \cite{park2019pix2pose111} adopts the idea of GANs to predict dense-coordinates for instance-level 6D object pose estimation.

 In this paper, we propose a novel and effective approach called ACR-Pose. Its Reconstructor taking use of two delicately designed submodules: PIM and RRM, to generate pose insensitive and essential relational information to get accurate reconstruction results. The Reconstructor and the Discriminator are optimized adversarially. As far as we know, we are the first to adopt adversarial training in canonical representation reconstruction.

\section{ACR-Pose}
In this section, we introduce ACR-Pose in detail. We first illustrate the overview of our pipeline (Sec. \ref{sec:overview}). Then we introduce at length the Pose-Irrelevant Module (Sec. \ref{sec:rot}) and Relational Reconstruction Module (Sec. \ref{sec:relation}) respectively. Next, we shortly depict the Shape Prior Deformation step \cite{tian2020shape} (Sec. \ref{sec:spd}) we used for reconstruction. After that, we describe how the Discriminator and adversarial training (Sec. \ref{sec:gan}) improve the reconstruction reality. Finally, we present the loss function.

\subsection{Pipeline}
\label{sec:overview}

In this paper, we aim to reconstruct the 3D canonical representation, for example, NOCS \cite{wang2019normalized} representation from a RGBD input. Fig. \ref{fig:trirpose} illustrates the overview of our pipeline. It can be divided into three parts.

The first part is the initialization phase. A detector is used to get the image patch and depth region of the target object from the RGBD input. The image patch and depth region are the main input of the following networks. Similar to \cite{tian2020shape}, we train an encoder-decoder network to learn a shape prior for each category and take the learned shape prior as one of the following network's input. Totally, we have three modalities of input.

 The second part is the Reconstructor. In this part, the image patch is fed into an Image Encoder backbone to learn instance RGB image features $f_I \in R^{U \times V \times C}$, where $U$ and $V$ are the image size. The depth region is back-projected into a point cloud (observed points) and fed into the \emph{PIM} to learn canonical-related instance point features $f_P \in R^{N_p \times C}$, where $N_p$ is the number of observed points. The shape prior is fed into a simple Shape Prior Encoder to learn high-dimensional category prior features $f_c \in R^{N_c \times C}$. Then, learned features of the three modalities are integrated and enhanced by the \emph{RRM}, which is another key component of our work. Finally, taking features learned by both modules as input, a Shape Prior Deformation \cite{tian2020shape} step is employed to reconstruct the NOCS representation.
 
 In the Discriminator, the third part, we utilize an adversarial scheme during training. Specifically, the reconstructed NOCS representation and the ground-truth are fed into a Discriminator. The Discriminator would then judge the realistic level of the reconstructed NOCS presentations as well as the ground-truth, hence guiding and encouraging our Reconstructor to generate NOCS representations as realistic as possible. Thus, the Reconstructor and the Discriminator would compete with each other to become stronger. We call it \emph{Adversarial Reconstruction}.
 
  At inference time, given the reconstructed NOCS representation and the back-projected observed object depth, we use the Umeyama algorithm \cite{umeyama1991least}  to recover the 6D object pose. Next, we introduce the PIM and the RRM in the Reconstructor first.

\begin{figure}[t]
  \centering
   \includegraphics[width=0.95\linewidth]{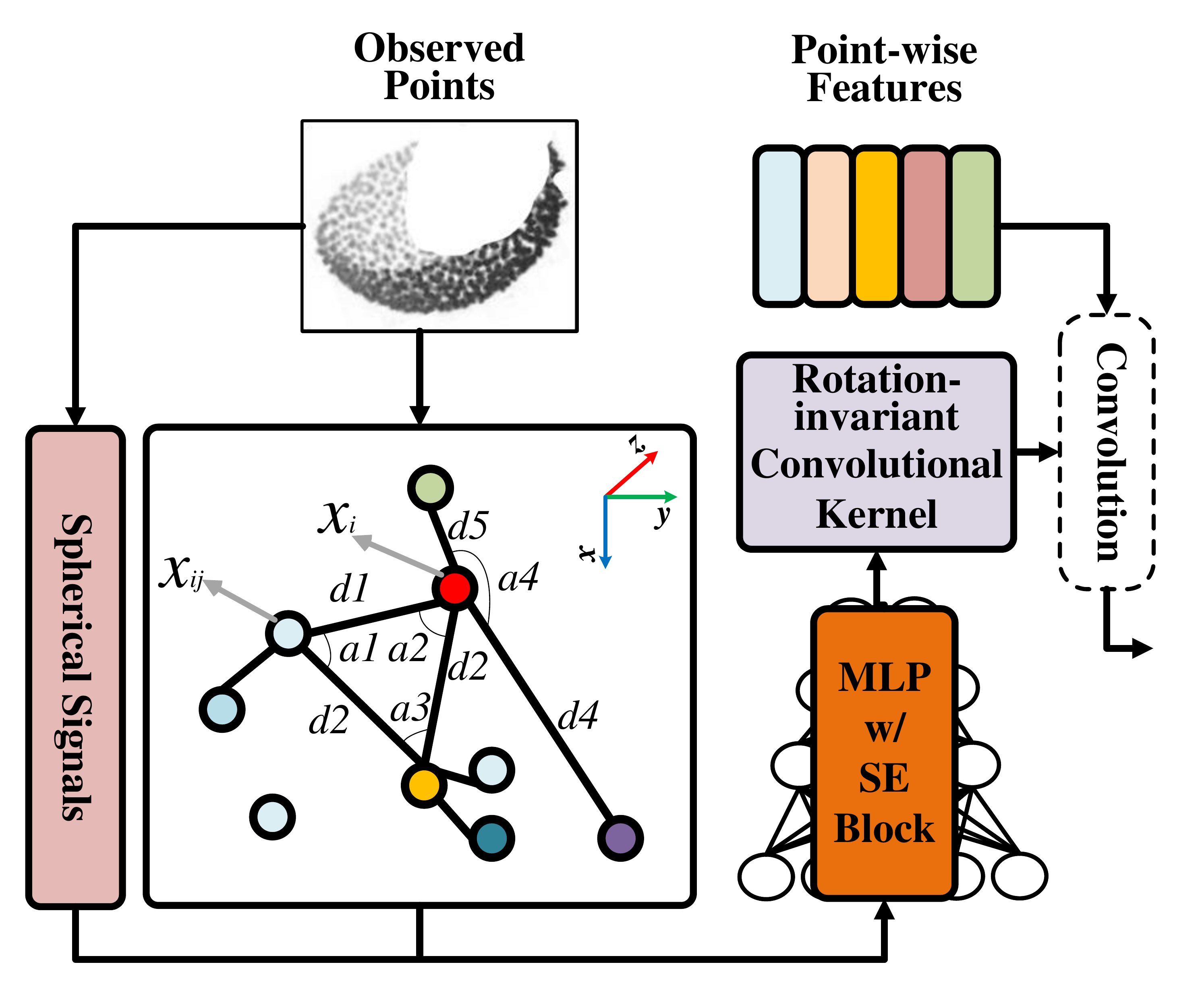}
   \vspace{-0.1in}
   \caption{Rotation-Invariant Convolution. Each point $x_i$ and its $K$ nearest points form a local group. First, we compute the mean geometric center (the yellow point) of the group, the nearest point of $x_i$ (the green point), and the farthest point of $x_i$ (the purple point). Then, for each point $x_{ij}$ in this group, nine low-level rotation-invariant geometry features (angles $a1$ to $a4$ and distances $d1$ to $d5$) and two spherical signals are calculated and utilized to learn the rotation-invariant convolutional kernels. Finally, a PointConv operation is employed to apply the learned kernels to the input point-wise features for feature updating.}
   \label{fig:rotconv}
    \vspace{-0.2in}
\end{figure}

\subsection{Pose-Irrelevant Module in Reconstructor}
\label{sec:rot}
The Reconstructor takes the back-projected object depth as one of its main inputs.
A normalization filter is occupied to eliminate the impact of translation and a Rotation-Invariant Network is utilized to filter the impact of rotation. These two modules together comprise the PIM.

Specifically, denoting the back-projected point cloud of the object as $P$, we first filter out the impact of the translation by subtracting its geometric center, i.e. $\hat{P}=P-\overline{P}$, where $\overline{P}$ is the mean geometric center. Then, we input $\hat{P}$ into 4 layers of Rotation-Invariant Convolution for learning rotation-invariant features, called canonical-related features. The Rotation-Invariant  Convolution is illustrated in Fig. \ref{fig:rotconv}. Following \cite{fan2021attentive}, given the input point cloud and its corresponding point-wise features as input, we choose nine low-level rotation-invariant geometry features (see Fig. \ref{fig:rotconv} middle) as well as two spherical signals \cite{you2021prin} to learn rotation-invariant convolutional kernels using a MLP and a SE-block \cite{hu2018squeeze}. Then we use a PointConv \cite{wu2019pointconv}  to apply learned kernels to  point cloud features for feature update.

The output of the PIM is $f_P \in R^{N_p \times C}$, where $N_p$ is the number of observed points in the point cloud or the depth region. It will be incorporated with features of the other two modalities to achieve the NOCS representation reconstruction task.

\subsection{Relational Reconstruction Module in Reconstructor}
\label{sec:relation}
As mentioned before, the Reconstructor takes three modalities of data as input, they all contribute to reconstructing the NOCS representation. Therefore, to learn high-quality features for high-quality NOCS representation reconstruction, we propose the RRM, which consists of three different graph convolutional networks to learn and integrate three kinds of  relational features in the feature space.  Details of the three kinds of relational features are below:

\textbf{Relational instance feature} is the point-wise feature of the observed object, including the texture feature and the geometry feature. To learn it, we first index each point's point feature and RGB image feature from $f_P$ and $f_I$, and concatenate them to form a $N_p \times 2C$  features matrix. A MLP is used to reduce the feature dimension to $N_p \times C$. Then, we construct a graph $G(V,E)$ in the feature space using the feature matrix, where  $V$ represents nodes in the graph, a node means an observed point, and $E$ represents edges. The adjacent relation in the graph is built using the K-Nearest-Neighbors (KNNs) algorithm. The KNN algorithm would first compute the pair-wise distance in the feature space for each point pair. Then, it chooses each point's $K$ nearest neighbors to build adjacent edges. Next, an  EdgeConv \cite{wang2019dynamic} operation is adopted to learn features taking the graph as input. Finally, we use another MLP to increase the per-point feature dimension back to $2C$. We denote the relational instance feature as $f_{ins} \in R^{N_p \times 2C}$. 

\textbf{Relational deformation feature} is the second kind of feature our network learns. It is mainly used  to deform the shape prior into a canonical instance model of the object in the Shape Prior Deformation step (detailed in Sec \ref{sec:spd}). To achieve this goal, we should make the feature be able to implicitly reflect the relational context between the shape prior and the observed object. Therefore, as before, we  use a graph network to learn this kind of relational information. Specifically, we use a MLP and an adaptive average pooling operation to embed the relational instance feature $f_{ins}$ into a global feature vector $v_{ins}$. The global prior feature vector $v_{c}$ is obtained in the same way. Then, we repeat $v_{ins}$ and   $v_{c}$  for $N_c$ times and  concatenate them with the prior features $f_c$. The concatenated features are fed into another MLP for dimension reduction.
After that, we build a graph  utilizing the output and pass messages between nodes using an EdgeConv. The final output is the relational deformation feature. We denote it as $f_d \in R^{N_c \times C} $.

\textbf{Relational assignment feature} is the third kind of feature we learn. In the Shape Prior Deformation step, we need to assign the above-mentioned deformed shape prior to the NOCS representation (detailed in Sec \ref{sec:spd}).  Since the deformed shape prior and the  NOCS representation are from different parameter fields, we need to establish a one-to-many correspondence (an assignment matrix) between them. To build the correspondence, we should mainly take advantages of two kinds of data, i.e., the observed instance points  and  the  shape variation. The former works for providing instance characteristics, and the latter works for reducing shape variations. To better utilize their features and flowing information between them,  we repeat $v_{ins}$ and   $v_{c}$  for $N_p$ times and concatenate them with the relational instance feature $f_{ins}$. Then, we use a graph network as before to learn the assignment feature $f_{a} \in R^{N_p \times C}$.

The \emph{Relational Reconstruction Module} uses graphs to learn high-quality features by integrating relational information between features of different modalities. Constructing graphs in the feature space is quite beneficial in learning more semantics about the object shape and category specialities. The learned features will be further used in the  Shape Prior Deformation step for the final reconstruction (Sec. \ref{sec:spd}), which would all contribute greatly to high-quality NOCS representation reconstruction, verified by experimental analysis.

\subsection{Shape Prior Deformation in Reconstructor}
\label{sec:spd}
After all necessary features are learned. We adopt the Shape Prior Deformation approach to the reconstruct the NOCS representation in the Reconstructor. Shape Prior Deformation is first proposed by Tian et al. \cite{tian2020shape}.  For the integrity of the paper, we briefly introduce it here.

The Shape Prior Deformation step requires us to learn two matrices to transform the shape prior into the NOCS representation.  They are the deformation field $M_d \in R^{N_c \times 3}$ and the assignment matrix  $M_a \in R^{N_p \times N_c}$. We feed the relational deformation feature into a MLP for learning $M_d$ and use another MLP to take the relational assignment features as input to learn $M_a$. The NOCS representation can be reconstructed by:

\begin{equation}
P_{nocs}={M_a}(P_c+M_d)
\end{equation}

where $P_c$ is the shape prior. Using the two-stage transformation operation, $P_c$ would first be transformed into the observed points' canonical instance model and then be transformed to the NOCS representation. The Shape Prior Deformation step ensures the robustness of the reconstruction.   It serves as the final step of our Reconstructor.

\begin{table*}[t]
		\centering
		\begin{tabular}{c |c|c|c|c|c|c|c }  
			\hline 	
			Data& Methods & IoU50& IoU75& 5\degree 2cm& 5\degree 5cm& 10\degree 2cm& 10\degree 5cm\\
			\hline 	
			\multirow{4}*{CAMERA} &NOCS\cite{wang2019normalized} &83.9 & 69.5& 32.3& 40.9& 48.2&64.6  \\
			&SPD\cite{tian2020shape} &93.2 & 83.1& 54.3&59.0& 73.3&81.5     \\
			&DualPoseNet\cite{lin2021dualposenet}&92.4 &86.4& 64.7&70.7&77.2&84.7 \\
			&FS-Net\cite{chen2021fs} &- & 85.2& -&62.0&-&60.8\\
			&ACR-Pose(Ours)&\textbf{93.8}&	\textbf{89.9}	&\textbf{70.4}&	\textbf{74.1}&	\textbf{82.6}&	\textbf{87.8}
 \\
					
			\hline 	
	\end{tabular}
	\vspace{-0.1in}
	\caption{Performance on NOCS-CAMERA dataset.}
	\label{camera}
\end{table*}

\begin{table*}[t]
		\centering
		\begin{tabular}{c |c|c|c|c|c|c|c }  
			\hline 	
			Data& Methods & IoU50& IoU75& 5\degree 2cm& 5\degree 5cm& 10\degree 2cm& 10\degree 5cm\\	
			
			\hline 	
			\multirow{6}*{REAL} &NOCS\cite{wang2019normalized}& 78.0 & 30.1& 7.2& 10.0& 13.8&25.2    \\
			&CASS\cite{chen2020learning}	& 77.7 & - & - &23.5& -&58.0 \\
			&SPD\cite{tian2020shape} 	&77.3& 53.2 & 19.3& 21.4& 43.2& 54.1     \\

			&DualPoseNet\cite{lin2021dualposenet}&79.8 &62.2&29.3& 35.9& 50.0&\textbf{66.8} \\
			&FS-Net\cite{chen2021fs} &\textbf{92.2}& 63.5& -&28.2&-&60.8\\
			
			&ACR-Pose(Ours)&82.8&	\textbf{66.0}	&\textbf{31.6}&	\textbf{36.9}&	\textbf{54.8}&	65.9
\\
			\hline 	
	\end{tabular}
	\vspace{-0.1in}
	\caption{Performance on NOCS-REAL dataset.}
	\label{real}
	\vspace{-0.15in}
\end{table*}
\subsection{Discriminator and Adversarial Training}
\label{sec:gan}
Though the NOCS representation can be predicted by only using the  Reconstructor, we observe that some reconstruction results are quite unrealistic, which may severely affect the 6D object pose estimation accuracy. In this paper, we propose to use an adversarial reconstruction strategy to solve the problem. Specifically, we propose to use a Discriminator to guide the Reconstructor. The Discriminator in our work consists of a MLP, a global max-pooling and two fully connected layers. It takes the reconstructed NOCS representation and the real one as input at training time. And the output is a probability value distributed from 0-1, where a  higher value means a larger probability it regards the input point cloud as a real NOCS representation and vice versa. The optimization goal of the Discriminator is to minimize:
\begin{equation}
L_d=(\mathbb{D}(\hat{P_{nocs}})-1)^2+(\mathbb{D}(\overline{P_{nocs}}))^2
\end{equation}

where  $\overline{P_{nocs}}$ is the reconstructed NOCS representation and  $\hat{P_{nocs}}$ is the ground-truth.

 Correspondingly, the optimization goal of the Reconstructor is to minimize:
\begin{equation}
L_g=(\mathbb{D}(\overline{P_{nocs}})-1)^2
\end{equation}

During training, the Discriminator would try its best to judge the difference between real and fake, while the Reconstructor would try its best to confuse the Discriminator. In the fierce competition, both of them will gradually become stronger and stronger, hence increasing the reality of the reconstructed NOCS representation.

\subsection{Loss Function}
Except for the adversarial loss $L_d$ and $L_g$ described before, we use the smooth L1  loss $L_{corr}$  between the reconstructed NOCS representation and ground-truth to encourage better one-to-one correspondence. We use the chamfer distance loss $L_{cd}$ between the deformed shape prior and the object's canonical instance model to preserve appearance information. We also use a cross-entropy loss $L_{entro}$ to encourage peak distribution of the assignment matrix $M_a$ and a L2 regularization loss $L_{reg}$ on $M_d$ to avoid collapsing deformation. The final loss function is:
\begin{equation}
L=\gamma_1 L_d+\gamma_2 L_g+ \gamma_3 L_{corr}+\gamma_4 L_{cd}+\gamma_5 L_{entro}+\gamma_6 L_{reg} 
\end{equation}
where $\gamma_1$ to $\gamma_6$ are balance terms.

\section{Experiments}
\subsection{Datasets}
We evaluate our method on the NOCS-CAMERA dataset and NOCS-REAL dataset proposed by Wang et al.\cite{wang2019normalized}, which currently are the most widely used and authoritative datasets for benchmarking category-level 6D object pose estimation methods. The NOCS-CAMERA dataset is a synthetic dataset that contains 300K RGBD images (with 25K for evaluation) generated by rendering and compositing synthetic objects into real scenes. The NOCS-REAL dataset is a real-world dataset that contains
4.3K real-world RGBD images from 7 scenes for training, and 2.75K real-world RGBD images from 6 scenes for evaluation. Both datasets consist of six categories, i.e., bottle, bowl, camera, can, laptop and mug.

\subsection{Implementation Details}
We choose Mask-RCNN \cite{he2017mask} as the detector. The Image Encoder is a PSP-Net \cite{zhao2017pyramid} with a ResNet-18 \cite{he2016deep} backbone. The Shape Prior Encoder is a MLP. We first train our model for 50 epochs on the  NOCS-CAMERA train set and evaluate it on the NOCS-CAMERA val set.  Then, we fine-tune it on the NOCS-CAMERA train set and NOCS-REAL train set for 10 epochs. And we evaluate our fine-tuned model on the NOCS-REAL test set.  Note we also have to recover the object size, we simply use the average size of $P_c+M_d$ as the result following \cite{tian2020shape}.  More details are in the SuppMat. 

\begin{figure*}[t]
  \centering
   \includegraphics[width=0.95\linewidth]{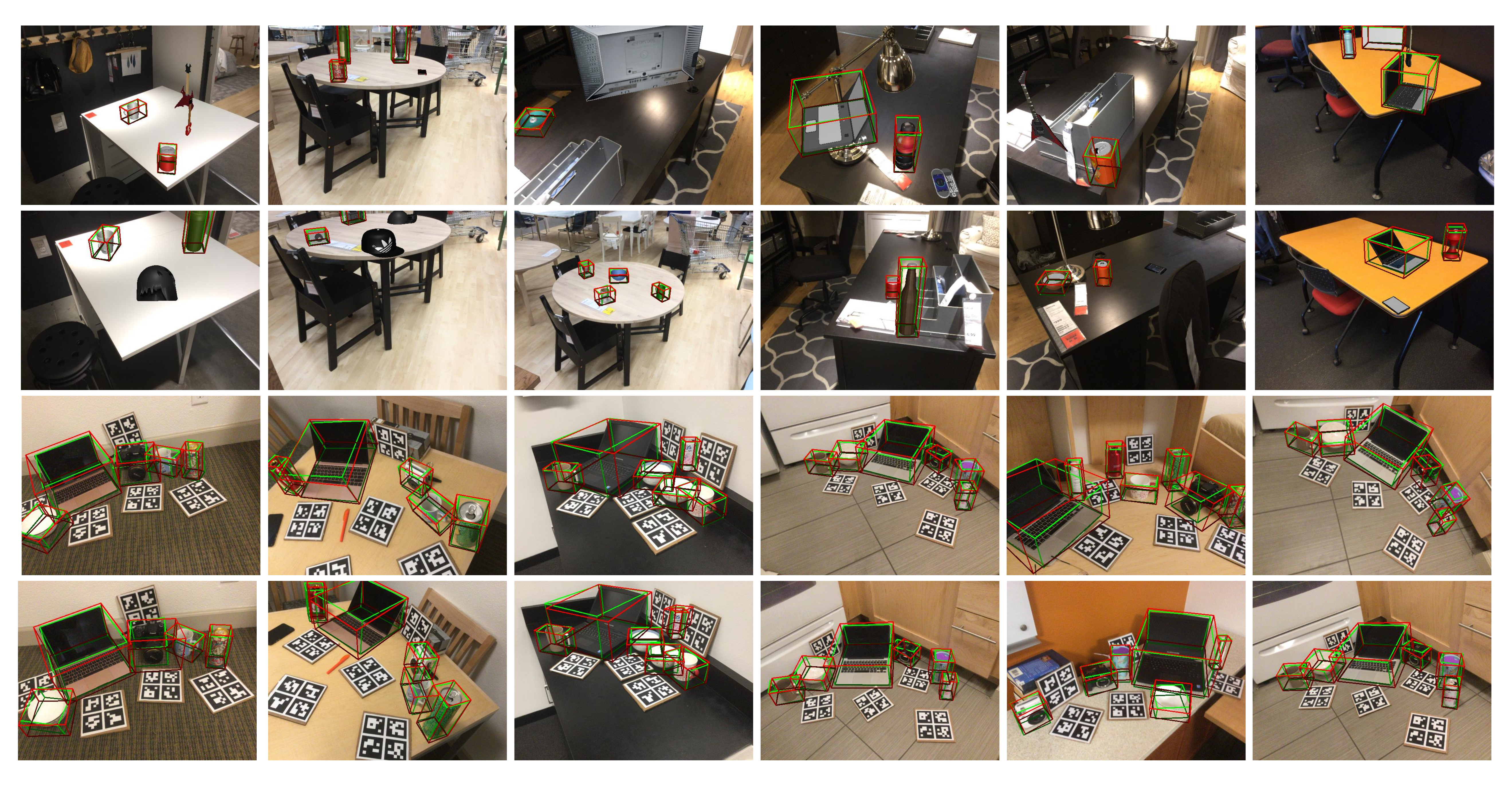}
   \vspace{-0.15in}
   \caption{Visualization of the prediction results of our method. Green boxes are the ground-truth and red boxes are our predictions. The top two rows are results on the NOCS-CAMERA dataset and the bottom two rows are results on the NOCS-REAL dataset.}
   \label{fig:visulizesucess}
\end{figure*}

\subsection{Comparison with State-of-the-art}
We compare the performance of ACR-Pose  with some recent state-of-the-art methods including NOCS \cite{wang2019normalized}, CASS \cite{chen2020learning}, SPD \cite{tian2020shape}, DualPoseNet \cite{lin2021dualposenet} and FS-Net \cite{chen2021fs}, to verify the effectiveness of our method.  All of them are strong baselines. For comparison, we compute the IoU50, IoU75, 5\degree 2cm, 5\degree 5cm, 10\degree 2cm and 10\degree 5cm metrics and report the mean Average Precision (mAP)  to measure the performance.
 
\textbf{Results on NOCS-CAMERA:} Table \ref{camera} shows the result of our method and the results of all competitors on the val set of NOCS-CAMERA. It is obvious that our method outperforms all the strong baselines for a large margin. For example, SPD \cite{tian2020shape} shares the same backbone and the same Shape Prior Deformation step as our work,  however, it is outperformed by our ACR-Pose by 16.3\%, 15.1\%, 9.3\% and 6.3\%  in terms of the 5\degree 2cm, 5\degree 5cm, 10\degree 2cm and 10\degree 5cm, respectively, which is a significant improvement. The comparison between SPD  \cite{tian2020shape} and our method demonstrates that the improvement really comes from our adversarial reconstruction scheme, which also benefits from the two proposed modules in the Reconstructor. They have increased our model's 6D object pose estimation performance by increasing the reconstruction quality and reality of the NOCS representations through adversarial training.  Our method also outperforms the second-best method DualPoseNet \cite{lin2021dualposenet} by 5.7\%, 3.4\%, 4.6\%, 3.1\% at the above four metrics respectively, which is also a great improvement. This further proves the effectiveness our method.

\begin{table*}[t]
		\centering
		\begin{tabular}{c|c|c|c|c|c|c }  
			\hline 	
			Versions & IoU50& IoU75& 5\degree 2cm& 5\degree 5cm& 10\degree 2cm& 10\degree 5cm\\	
			
			\hline 	
Baseline&	93.3	&86.7&	58.6	&63.3	&76.4	&84.2\\
+ Relational Instance Features&	93.8	&84.6	&62.4	&67.0	&78.8&	85.7\\
+ Pose Irrelevant Module	&93.4	&88.9	&65.5	&69.2&	79.7&	85.2\\
+ Adversarial Reconstruction&	93.7&	84.0&	67.4&	71.0&	81.2&	86.3\\
+ Relational Deformation Features&	93.7&89.3&	68.4&	72.1&	81.9&	87.0				\\	
+ Relational Assignment Features (Full model)&\textbf{93.8}&	\textbf{89.9	}&\textbf{70.4}&	\textbf{74.1}&	\textbf{82.6}	&\textbf{87.8}\\
			\hline 	
	\end{tabular}
	\vspace{-0.1in}
	\caption{Results of ablation study. We gradually add our designs one by one to the baseline to investigate their impacts.}
	\label{ablation}
	\vspace{-0.1in}
\end{table*}

\textbf{Results on NOCS-REAL}: We also compare ACR-Pose with other methods on the NOCS-REAL test set. Results are shown in Table \ref{real}. Our method is defeated by FS-Net \cite{chen2021fs} at the IoU50 metric and defeated by DualPoseNet \cite{lin2021dualposenet} at the 10\degree 5cm metric. However, these two metrics are relatively loose metrics. Besides, the training setting of FS-Net \cite{chen2021fs} is strongly different from all the other baselines. For example, it uses the YOLO v3 \cite{redmon2018yolov3} object detector to detect and crop image patches and depth regions,  while all the other baselines use the Mask-RCNN detector. When it comes to other strict metrics, our method outperforms other baselines for a large margin. The excellent performance of our method on the NOCS-REAL dataset at more strict metrics demonstrate that ACR-Pose is powered with strong generalization ability and owns the potential of being deployed into real-world industry products.  We contribute the superiority of our model into PIM, RRM and the adversarial reconstruction scheme.

\begin{figure}[t]
  \centering
   \includegraphics[width=0.98\linewidth]{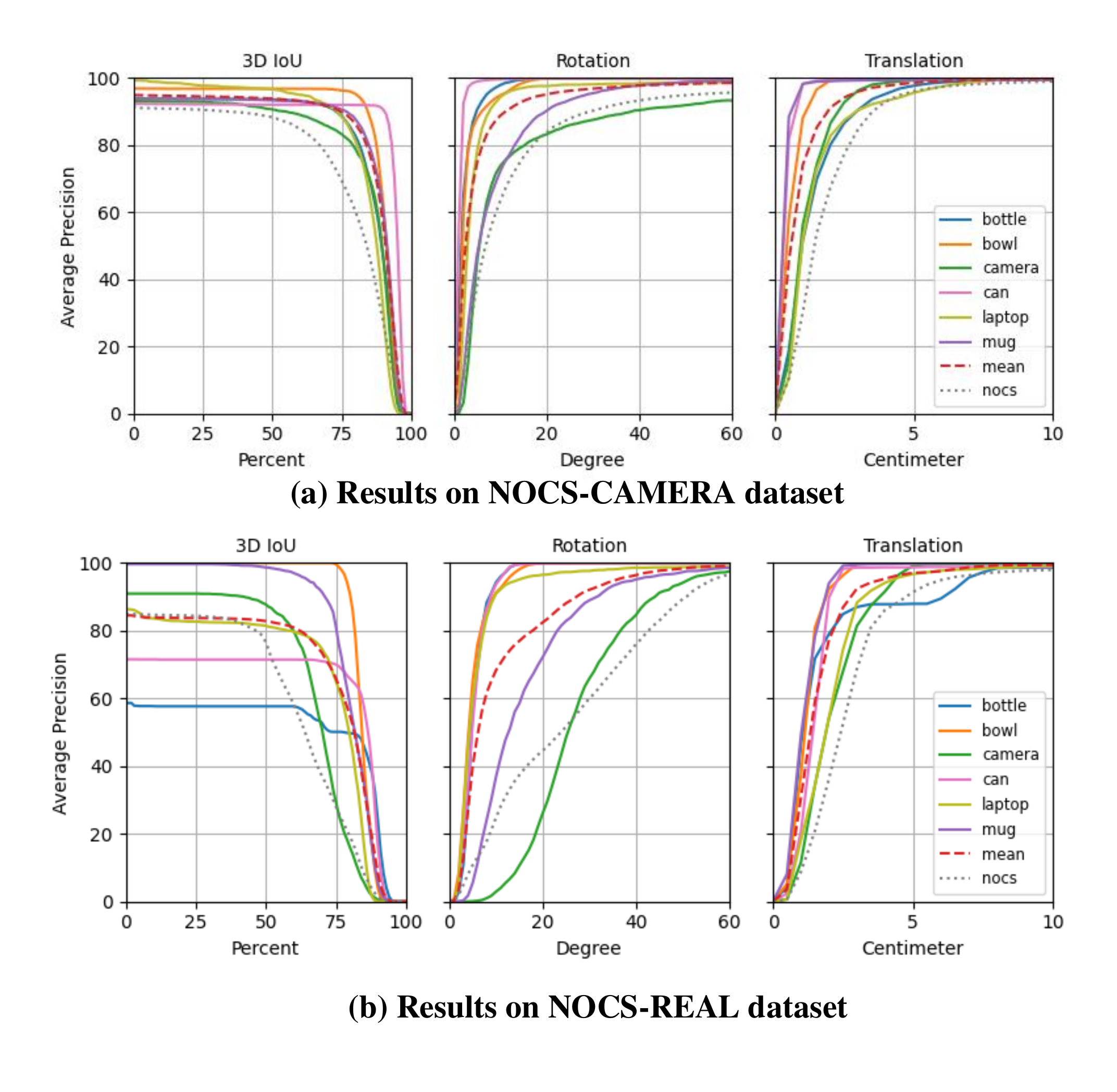}
    \vspace{-0.25in}
   \caption{The average precision (AP) vs. different thresholds on 3D IoU, rotation error, and translation error.}
   \label{fig:map_curve}
      \vspace{-0.25in}
\end{figure}

We also illustrate the average precision (AP) vs. different thresholds on 3D IoU, rotation error, and translation error of our method on Fig. \ref{fig:map_curve}. We compare our method with the mean result of \cite{wang2019dynamic}. Both versions of our method achieve excellent performance. And our method performs well for all categories.

\begin{figure}[t]
  \centering
   \includegraphics[width=0.95\linewidth]{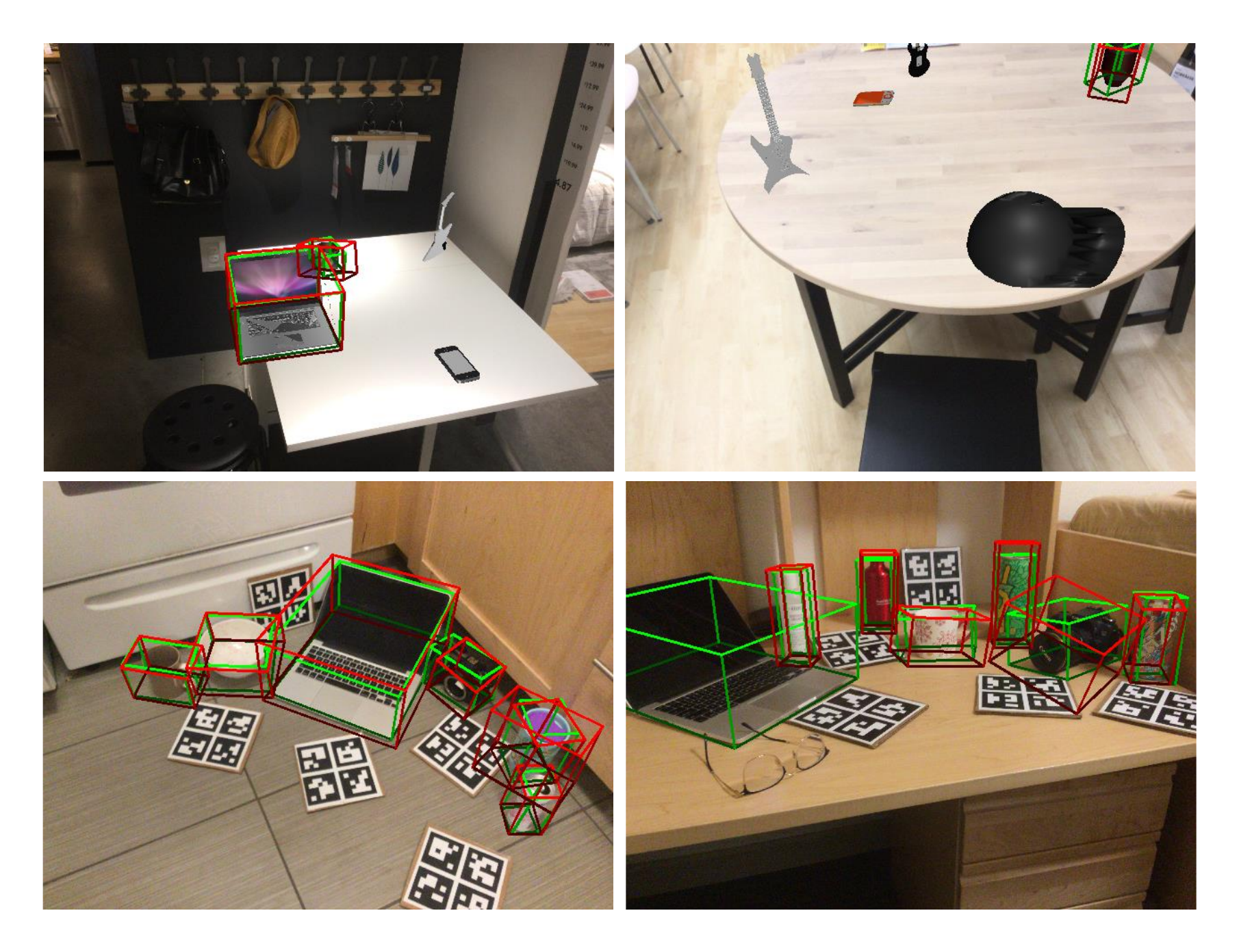}
   
   \vspace{-0.2in}
   \caption{Visualization of failure cases.}
   \label{fig:visulizefail}
   \vspace{-0.2in}
\end{figure}

\textbf{Qualitative results}: To better exhibit the performance of ACR-Pose, we visualize some prediction results on Fig. \ref{fig:visulizesucess}. It can be seen that our model can predict accurate 6D object pose because the visualized bounding boxes have high overlap with the ground-truth and they can frame the target objects tightly. Our method achieves excellent performance in both synthetic and real scenarios. Fig. \ref{fig:visulizefail} also illustrates some failure cases. When there exists truncation or occlusion, our model sometimes fails. And some objects may be missed by the detector. We will explore how to solve these limitations in our future work. More visualization results can be found in the SuppMat.

\subsection{Ablation Study}
In this section, we conduct experiments on the NOCS-CAMERA dataset to investigate the effectiveness and necessity of our several design choices.  In Table \ref{ablation}, we replace all of our novel network designs in ACR-Pose as MLPs and then gradually add them back one by one to study their impact on the final 6D object pose estimation results.

\textbf{Impact of the Relational Reconstruction Module:} The RRM is proposed to learn relational information between three different modalities in our Reconstructor. It includes three graphs that learn relational instance features, relational deformation features and relational assignment features respectively. From row 4, row 6 and row 7 of Table \ref{ablation}, we can find that all the three kinds of features contribute to the accuracy improvement because adding them one by one cumulatively increases the model's performance. The improvement comes from the message passing through graphs in the feature space. Using them, powerful semantic features can be learned by the model to better reconstruct the canonical representation. These semantics can effectively describe both  object-specified shape and  common category-level characteristics (category specialities), therefore,   high-quality NOCS representations can be reconstructed. The above results  demonstrate that relational features are essential for high-quality NOCS representation reconstruction.

\textbf{Impact of the Pose-Irrelevant Module:} The PIM contributes to learning canonical-related features to filter rotation-related and translation-related information that is irrelevant to high-quality NOCS representation reconstruction in the Reconstructor. When it is added to our baseline, all evaluation metrics are increased by  1\% to 3\%. That is because, without the interference of rotation and translation, the model can learn more information about the shape of the object's observed part, which is critical for reconstructing the NOCS representation. 

\textbf{Impact of Adversarial Reconstruction:} The adversarial reconstruction scheme plays a role of increasing reconstruction reality.  Besides, since the Discriminator is only required  at the training time,  it is an very efficient strategy for improving performance. The 5th row of Table \ref{ablation}  evidently verifies that the adversarial training scheme would help the model learn better, more realistic and more high-quality canonical representations because adding it brings significant  performance gains.  That is because the reconstructed canonical representations would be more realistic after adding the Discriminator.  Further, the following Umeyama algorithm \cite{umeyama1991least} would also perform  much better and more robust if it receives more realistic canonical representations.

\section{Conclusion}
In this paper, we present an Adversarial Canonical Representation Reconstruction Network, the ACR-Pose, for accurate category-level 6D object pose estimation. Our work focuses on reconstructing high-quality canonical representations for the observed objects, given the RGB patch, the depth region, and the learned shape prior. To achieve so, we propose an adversarial reconstruction scheme, which competitively trains a Reconstructor and a Discriminator to improve reconstruction performance. Besides, within our reconstructor, a Pose-Irrelevant Module is introduced to better extract the shape information of the objects, as well as a Relational Reconstruction Module that leverages the relational information among the three input sources. Experiments are conducted on synthetic and real datasets, where the effectiveness of each novel module is demonstrated, and our overall model achieves state-of-the-art performance.

One of the limitations of our method is that it may fail in the case of truncation or occlusion, which we aim to address as part of our future works.


{\small
\bibliographystyle{ieee_fullname}
\bibliography{egbib}
}

\section{Appendix}
\subsection{Implementation details}
ACR-Pose is implemented by PyTorch. The model is optimized by the Adam optimizer with a  batch size of 96. We choose Mask-RCNN \cite{he2017mask} as the detector. The Image Encoder is a PSP-Net \cite{zhao2017pyramid} with a ResNet-18 \cite{he2016deep} backbone. The Shape Prior Encoder is a MLP. The category instances are sampled from the ShapeNet dataset. The feature dimension C is set to 64. During training, we alternately update the parameters of the Reconstructor and the Discriminator.  The initial learning rates for training the Reconstructor and the Discriminator are 0.0001  and 0.00001 respectively. When training on the NOCS-CAMERA train set only, the learning rate is decayed at the 10th, 30th and 40th epoch. The decay rates are 0.5, 0.1, 0.01 w.r.t the initial learning rate repetitively.  When fine-tuning, the initial learning rates are also 0.0001  and 0.00001, and they are decayed by half at the 5th epoch. The image patch is resized to $192 \times 192$. The number of object points (back-projected object depth) $P$ is 1024. The number of shape prior points is also 1024. The number of adjacent neighbors  $K$ in the feature graphs is 36.  The balance terms  $\gamma_1$ to $\gamma_6$ in the loss function are 0.1, 0.1, 1.0, 5.0, 0.0001 and 0.01, respectively. When fine-tuning, we re-initial the discriminator and select data randomly from NOCS-CAMERA and NOCS-REAL at a ratio of 3:1 to train the network, following \cite{wang2019normalized,tian2020shape}.  All experiments are conducted on a single A6000 GPU.

\subsection{Overall accuracy}
Except for the mAP metric we mainly report, we also calculate the overall accuracy at the threshold of  IoU50, IoU75, 5\degree 2cm, 5\degree 5cm, 10\degree 2cm and 10\degree 5cm. We show the results on Table \ref{overallaccuracy} and compare our ACR-Pose with SPD \cite{tian2020shape}. Our method outperforms SPD for a large margin in terms of overall accuracy. This is consistent with the result of using mAP as an evaluation metric, which further indicates the superiority of our method.

\subsection{Generalization ability evaluation}
To evaluate the generalization ability, we evaluate ACR-Pose on the NOCA-REAL test set using the model only trained on the synthetic NOCS-CAMERA dataset. We also compare our method with SPD here. The evaluation metric is mAP. Results are shown in Table \ref{generalization}. Our method demonstrates better generalization ability towards real-world scenarios compared to SPD. However, there is still much room for performance improvement compared to the model trained on both NOCS-CAMERA and NOCS-REAL.

\subsection{Reconstruction quality evaluation}
To verify whether our model can reconstruct high-quality NOCS representations, we evaluate the reconstruction quality of the results of SPD and ACR-Pose using the chamfer distance. Results are shown in Table \ref{reconstructionquality}. Our method significantly outperforms SPD. Since we use the same Shape Prior Deformation step and Umeyama algorithm for NOCS representation reconstruction and object pose estimation, we claim that the mAP improvement really comes from the higher quality NOCS representation. Note our work mainly benefits from the Pose-Irrelevant Module, the Relational Reconstruction Module and the adversarial reconstruction scheme, while SPD doesn't use these modules.

\subsection{Impact of adjacent neighbors}
For the Relational Reconstruction Module, in each graph, we use KNN to search for the adjacent neighbors for each node. In Table \ref{impact_k}, we show the impact of the number of adjacent neighbors K.  We conclude that setting K as 36 is the best choice. When K is too small, it may cause a small receptive field for learning relationships between different modalities. While when it is too large, it may cause over-fitting and additional unnecessary computational costs.

\subsection{More visualization results}
\begin{figure*}[t]
  \centering
   \includegraphics[width=0.95\linewidth]{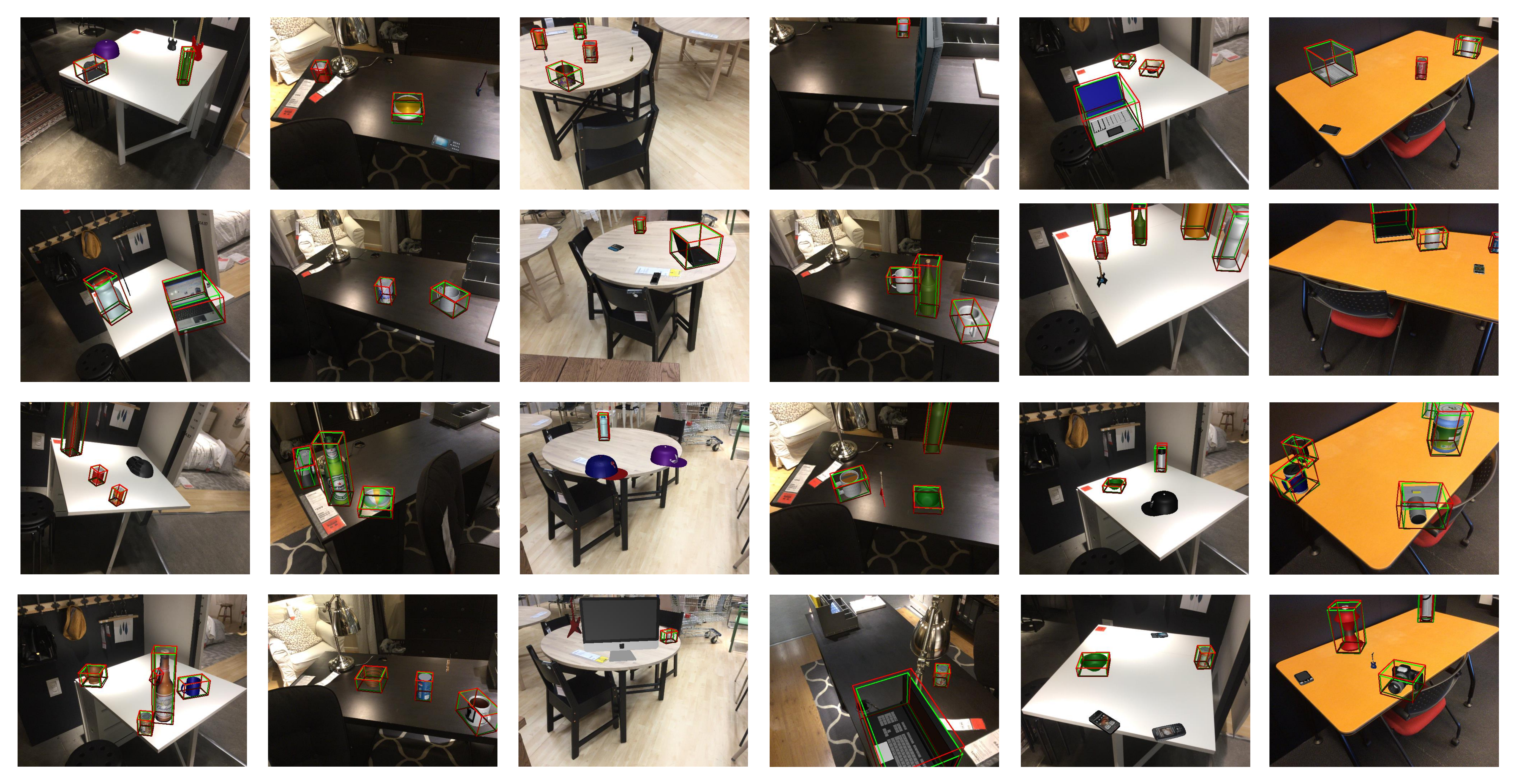}
   \caption{More visualization results on NOCA-CAMERA dataset.}
   \label{suppcamera}
\end{figure*}

\begin{figure*}[t]
  \centering
   \includegraphics[width=0.95\linewidth]{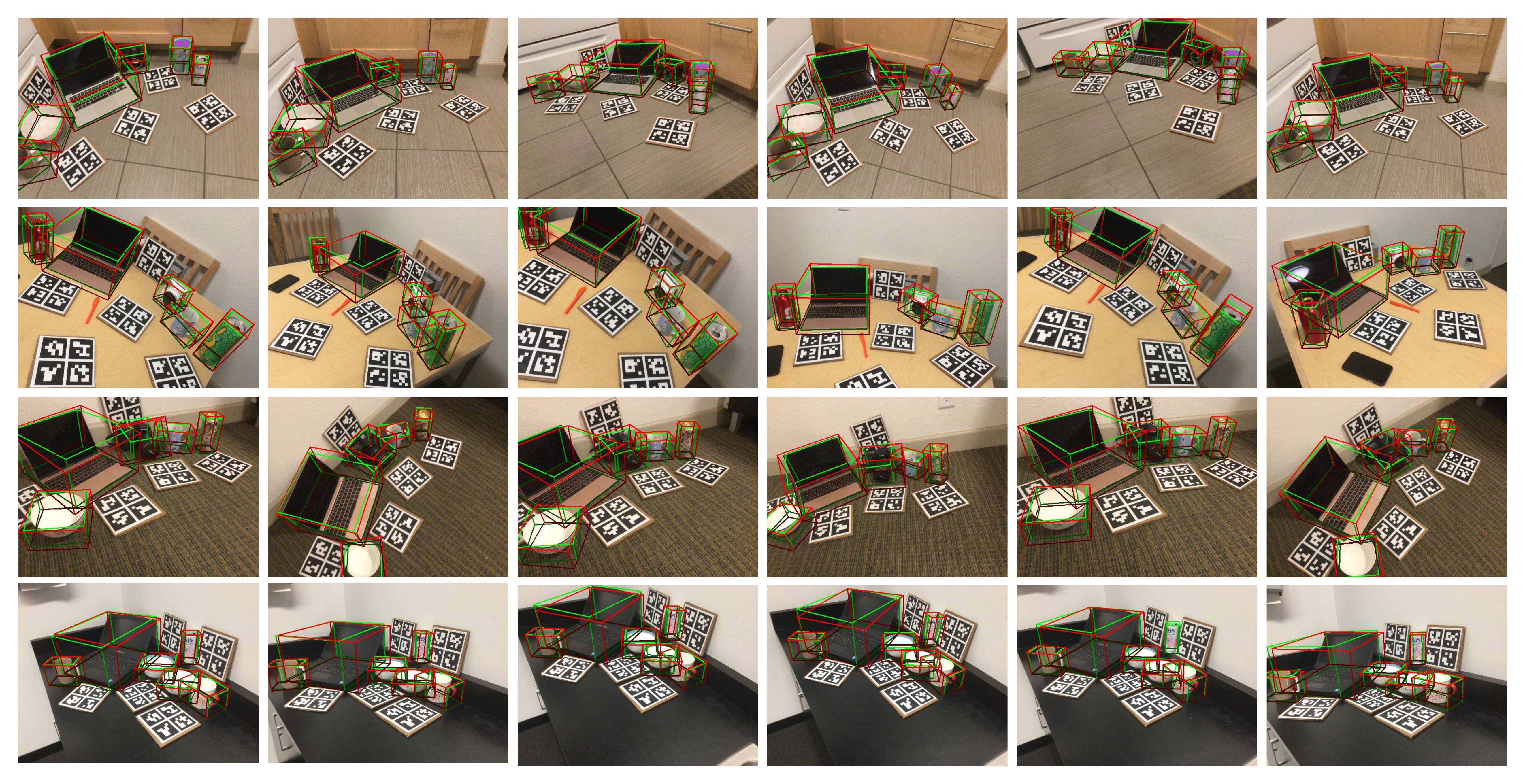}
   \caption{More visualization results on NOCA-REAL dataset.}
   \label{suppreal}
\end{figure*}

\begin{figure*}[t]
  \centering
   \includegraphics[width=0.95\linewidth]{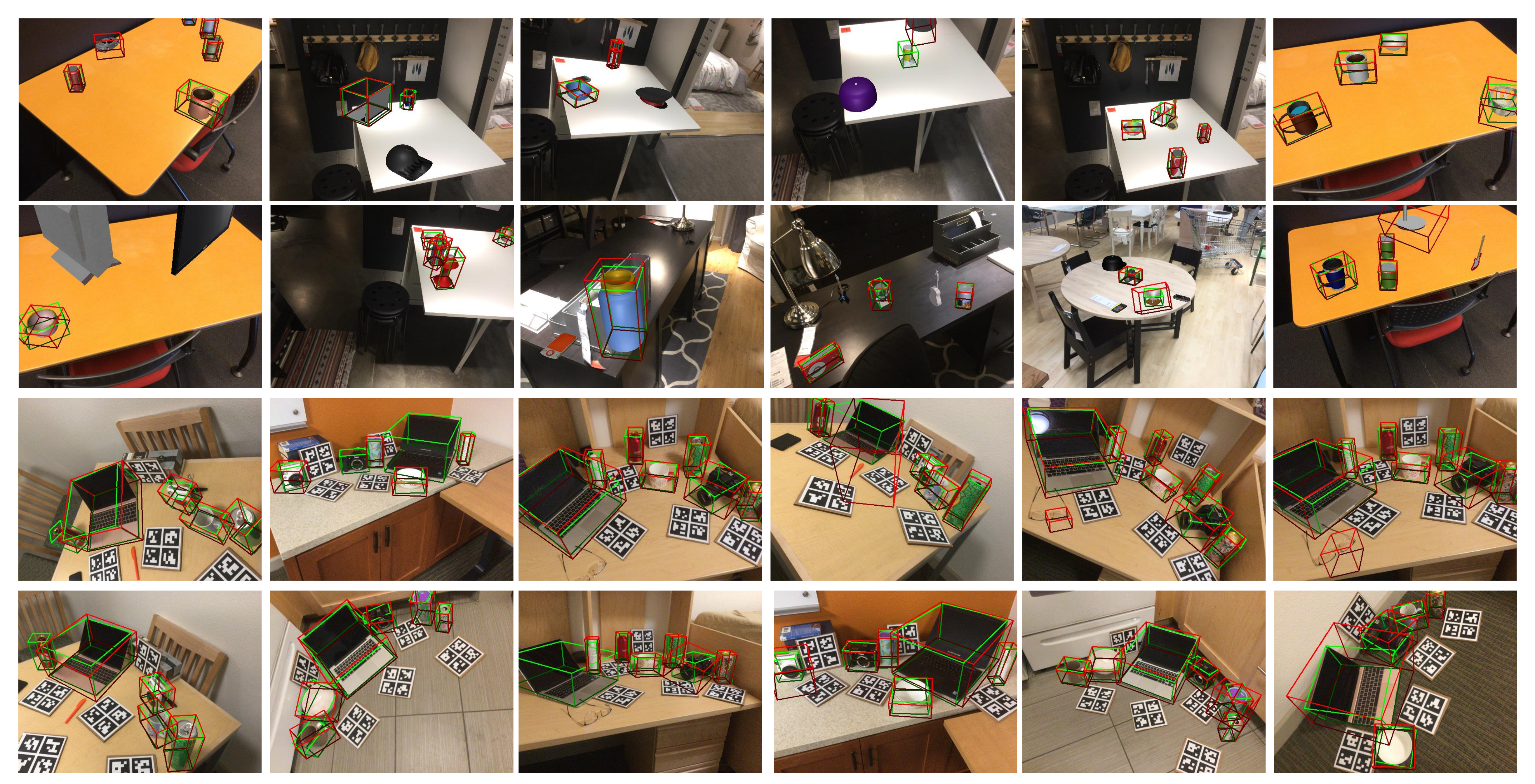}
   \caption{More visualization results of failure cases.}
   \label{suppfail}
\end{figure*}

To better understand the performance of our method. We illustrate more visualization results on Fig. \ref{suppcamera} and Fig. \ref{suppreal}. It can be seen that no matter in real-world scenarios or in synthetic scenarios, our model can accurately detect all the objects and recover their poses.  We also show more failure cases in Fig. \ref{suppfail}. Except for truncation and occlusion, our method may also sometimes suffers from ambiguities caused by symmetry. In our future work, we will investigate how to solve these disadvantages.

\begin{table*}[t]
		\centering
		\begin{tabular}{c |c|c|c|c|c|c|c }  
			\hline 	
			Data& Methods & IoU50& IoU75& 5\degree 2cm& 5\degree 5cm& 10\degree 2cm& 10\degree 5cm\\
	   \hline 
		\multirow{2}*{CAMERA}& SPD\cite{tian2020shape}&84.5&78.8 &67.5 &71.2 &82.3& 88.1  \\
			& ACR-Pose(Ours) &	\textbf{84.8}	&\textbf{82.6}	&\textbf{80.4}&	\textbf{83.0}	&\textbf{88.6}	&\textbf{92.1}\\
		\hline
		\multirow{2}*{Real}& SPD\cite{tian2020shape} & 87.8&69.4&31.4 &34.4&54.9 &63.1   \\
			& ACR-Pose(Ours) &\textbf{91.1}&	\textbf{79.2}&	\textbf{44.8}&	\textbf{49.9}&	\textbf{65.0}&	\textbf{73.4}\\
     \hline

	\end{tabular}
	\caption{Performance of SPD and our method  evaluated by the overall accuracy metric.}
	\label{overallaccuracy}
\end{table*}

\begin{table*}[t]
		\centering
		\begin{tabular}{c |c|c|c|c|c|c }  
			\hline 	
			Methods & IoU50& IoU75& 5\degree 2cm& 5\degree 5cm& 10\degree 2cm& 10\degree 5cm\\
	   \hline 
		 SPD\cite{tian2020shape} CAMERA only)&70.0 &\textbf{37.9} &11.2 &12.9&32.2 &41.3   \\
			ACR-Pose(CAMERA only)&\textbf{70.0}&35.8	&\textbf{12.2} &\textbf{14.8}&\textbf{32.2}	&\textbf{42.9}\\
			ACR-Pose(CAMERA+REAL) &\textbf{82.8}&	\textbf{66.0}	&\textbf{31.6}&	\textbf{36.9}&	\textbf{54.8}&	\textbf{65.9}\\

\hline 

	\end{tabular}
	\caption{Results of generalization ability evaluation.}
	\label{generalization}
\end{table*}

\begin{table*}[t]
		\centering
		\begin{tabular}{c |c|c|c|c|c|c|c }  
			\hline 	
			Data& Methods & bottle& bowl& camera& can& laptop& mug\\
	   \hline 
		\multirow{2}*{CAMERA}& SPD\cite{tian2020shape}&0.0235& 0.0142&0.0196 &0.0262 &0.0153&0.0187  \\
			& ACR-Pose(Ours) &	\textbf{0.0216}	&\textbf{0.0137}	&\textbf{0.0185}&	\textbf{0.0253}	&\textbf{
0.0137}	&\textbf{0.0172}\\
		\hline
		\multirow{2}*{Real}& SPD\cite{tian2020shape} &0.0251 &0.0192& 
0.0209&0.0213&0.0207 & 0.0210  \\
			& ACR-Pose(Ours) &	\textbf{0.0235}	&\textbf{0.0157}	&\textbf{0.0196}&	\textbf{0.0182}	&\textbf{0.0180}	&\textbf{0.0175}\\
     \hline 

	\end{tabular}
	\caption{Reconstruction quality evaluation. The evaluation metric is chamfer distance.}
	\label{reconstructionquality}
\end{table*}

\begin{table*}[t]
		\centering
		\begin{tabular}{c|c|c|c|c|c|c }  
			\hline 	
			Versions &IoU50&IoU75&5\degree 2cm& 5\degree 5cm& 10\degree 2cm& 10\degree 5cm\\	
			
			\hline 	
       K=24	& 94.5&93.7&69.9	&73.7	&82.2	&87.5\\
       K=36	& \textbf{94.5}&\textbf{93.8}&\textbf{70.4}	&\textbf{74.1}	&\textbf{82.6}	&\textbf{87.8}\\
       K=48	& 94.4&93.7&70.1	&73.7	&82.8	&87.8\\

			\hline 	
	\end{tabular}
	\caption{Impact of the number of adjacency neighbors.}
	\label{impact_k}
\end{table*}	

\end{document}